# The EcoLexicon English Corpus as an Open Corpus in Sketch Engine

*Pilar León-Araúz[1], Antonio San Martín[2], Arianne Reimerink[1]*
[1]*Department of Translation and Interpreting, University of Granada*
[2]*Department of Modern Languages and Translation, University of Quebec in Trois-Rivières*
*E-mail: pleon@ugr.es, antonio.san.martin.pizarro@uqtr.ca, arianne@ugr.es*

**Abstract**

The EcoLexicon English Corpus (EEC) is a 23.1-million-word corpus of contemporary environmental texts. It was compiled by the LexiCon research group for the development of EcoLexicon (Faber, León-Araúz & Reimerink 2016; San Martín et al. 2017), a terminological knowledge base on the environment. It is available as an open corpus in the well-known corpus query system Sketch Engine (Kilgarriff et al. 2014), which means that any user, even without a subscription, can freely access and query the corpus. In this paper, the EEC is introduced by describing how it was built and compiled and how it can be queried and exploited, based both on the functionalities provided by Sketch Engine and on the parameters in which the texts in the EEC are classified.

**Keywords:** specialized open corpus, terminology, corpus exploitation

## 1    Introduction

Corpora have become a key element of almost all language studies, as any assertion about language requires verification through real linguistic data to be deemed credible (Teubert 2005: 1). Having access to general and specialized corpora is thus essential for anyone involved in research or any professional activity related to language. However, many of these professionals do not have time to compile large corpora. The EcoLexicon English Corpus (EEC) is a 23.1-million-word specialized corpus of contemporary environmental texts. It was compiled by the LexiCon research group for the development of EcoLexicon (Faber et al. 2016; San Martín et al. 2017), a terminological knowledge base on the environment.[1] In EcoLexicon, the EEC and its Spanish counterpart (together over 50 million words) can be queried with pragmatic restrictions such as author, date of publication, target reader, contextual domain, and keywords. However, its search engine does not provide all the functionalities of the well-known corpus tool Sketch Engine (Kilgarriff et al. 2014). This is why the EEC was made available as an open corpus in Sketch Engine, which means that any user, even without a subscription, can freely access and query the corpus.[2] One very interesting module provided by the query system is information extraction through word sketches, which are automatic corpus-derived summaries of a word's grammatical and collocational behavior (Kilgarriff et al. 2010). Apart from the built-in word sketches, Sketch Engine allows users to customize sketches for their specific needs. In the case of the EEC, this has enhanced the extraction of semantic information.

In this paper, the EEC is introduced by describing how the corpus was built and compiled (Section 2), and how it can be queried and exploited (Section 3), based on the functionalities provided by Sketch Engine, the parameters in which the texts in the EEC are classified and the word sketches exclusively created for the EEC. Finally, Section 4 offers some concluding remarks.

---

1    EcoLexicon is freely available at <http://ecolexicon.ugr.es>.
2    Certain advanced functionalities are only available for subscribed users.

## 2    Creating the EcoLexicon English Corpus

The EEC is a 23.1-million-word corpus of contemporary environmental texts. It was first created as an internal tool for knowledge extraction while building EcoLexicon. However, it was made publicly available because it evolved to be a tool in itself that terminologists, translators or even experts could exploit for different purposes (i.e. modeling, comprehension and production tasks) within the specialized domain of the environment. As Sinclair (1991: 24) pointed out, we should not expect a general reference corpus like the British National Corpus to adequately document specialized genres and domains. It follows that we need more specialized corpora, compiled with enough texts and text types to represent a knowledge domain, as they are more likely to document the conventions of the genre and the concepts and terms of the domain.

Each text in the EEC is tagged according to a set of XML-based metadata, some of which are based on the Dublin Core Schema, while others have been included to meet the needs of the research group. Corpus metadata permit users to constrain corpus queries based on pragmatic factors, such as environmental domains and target reader. Thus, for instance, the use of the same term in different contexts can be compared. Tags are based on the following main parameters:

- Domain: the EEC encompasses all the domains and subdomains of environmental studies (e.g., Biology, Meteorology, Ecology, Environmental Engineering, Environmental Law, etc.).
- User: the corpus includes texts for three types of user, depending on level of expertise (i.e., expert, semi-expert, general public).
- Geographical variant: it comprises American, British, and Euro English.
- Genre: it covers a wide variety of text genres (e.g., journal articles, books, websites, lexicographical material, etc.).
- Editor: it distinguishes texts edited by scholars/researchers, businesses, government bodies, etc.
- Year: it includes texts from 1973 to 2016.
- Country: the texts are tagged according to the country of publication.

The EEC was processed and compiled in an internal application of the research group. Then it was recompiled within Sketch Engine with the Penn Treebank tagset (TreeTagger version 3.3) and with the EcoLexicon Semantic Sketch Grammar (ESSG) (León-Araúz & San Martín 2018; León-Araúz, San Martín & Faber 2016), a CQL-based (Corpus Query Language) (Jakubíček et al. 2010) customized sketch grammar separate from the default sketch grammar. The ESSG was developed for the extraction of semantic word sketches based on some of the most common semantic relations in terminology: generic-specific, part-whole, location, cause, and function.

When a corpus is compiled with a collection of different pattern-based grammar rules such as the above, new word sketches can be queried within the Sketch Engine (see Section 3.2). The ESSG thus has three aims: (1) extracting semantic relations for building EcoLexicon; (2) offering semantic word sketches in the EEC; and (3) providing other users with the possibility of reusing them in their own corpora.[3]

## 3    Exploiting the EcoLexicon English Corpus

The combination of pragmatic, syntactic and semantic information that can be extracted from the corpus makes the EEC an adequate resource for all kinds of end users with an interest in environmental science, such as domain experts, professional writers, translators, terminologists, ESP researchers, etc., as stated above. Thanks to Sketch Engine's automation capabilities, users are able to analyze and extract a sizable quantity of linguistic data that would have been unmanageable in the past (Kosem et al. 2014: 362). In the following sections, different queries will be provided by combining the main functionalities of Sketch Engine with the parameters according to which the EEC is tagged.[4]

---

3    The latest version of the ESSG can be downloaded from <http://ecolexicon.ugr.es/essg/>.

### 3.1   Search and Text Types

The feature *Search* is the main way to access concordances in Sketch Engine. Different types of queries are possible (simple, lemma, phrase, word, character and CQL), and they can be combined with the contextual filter, which allows the user to limit the lemmas that should appear around the word or words of the query. Additionally, in the case of the EEC, any query performed through the *Search* feature can be filtered according to text type based on the tagging of the EEC (domain, genre, editor, etc.) (Figure 1).

The filtering by text type can be chosen manually for each query. However, the user can also create subcorpora based on text types. For instance, a user may want to create a simple subcorpus for the domains of Hydrology or Renewable Energy, or complex subcorpora, such as one containing only articles and books in British English from the domain of Biology for experts in the field. Additionally, the EEC comes with several subcorpora created by default (i.e. American English, British English, Year 1973–1999, Year 2000–2009 and Year 2010–2016).

Figure 1: Sketch Engine's Search and EEC Text types.

All these possibilities of query customization allow the user to retrieve, for instance, all the concordances where *recycle* is a verb in texts addressed to the general public (lemma search filtered by user) or where *climate change* occurs in Environmental Law texts (phrase search filtered by domain). Additionally, the *Context* option can be combined with any search, permitting the user to find, for example, all the concordances in Oceanography academic articles where the lemma *wind* appears in a window of ±15 tokens of the lemma *wave*.

However, given that the EEC was recompiled with TreeTagger, it is possible to perform more fine-grained queries in CQL, allowing for the formalization of grammar patterns in the form of regular expressions combined with POS-tags. CQL queries used together with text-type filtering are a powerful tool to research the workings of environmental English. An example of a CQL query is *([tag="N.*"] [lemma="amount" & tag="N.*"])|([lemma="amount" & tag="N.*"] [word="of"] [tag="N.*"])*, which finds concordances of the lemma *amount* either preceded by any noun or followed by *of* and any noun. Figure 2 shows a sample of the resulting concordances limited to the Meteorology subdomain.

---

4   Due to space restrictions, no instructions are provided. However, interested readers can consult the user-friendly Sketch Engine manual at: < http://sketchengine.co.uk/user-guide/>

| | | |
|---|---|---|
| fall centers in June. However , the simulated | rainfall amount | and distribution in June in the uncoupled mo |
| uclear power plants do not lead to significant | amounts of contamination | , potential accidents in the nuclear industry |
| and some extreme dust events bring the same | amount of dust | in a few hours as the cumulated amount ove |
| chetti et al. , 2008 ). Such fires produce large | amounts of gases | and particles from biomass burning and soil |
| oles and the loss of cross-correlation between | precipitation amounts | and other weather variables ( [ Rajagopalan |
| e the plots of the right panel ) show that high | precipitation amounts | are derived from the following specific weat |
| the observed wind directions ; however , the | amount of wind | from the north and north-east is slightly und |
| cooling of air. Warm air is able to hold larger | amounts of water | vapor than cool air , so when air cools it is n |
| und , gaining temperature and increasing the | amount of water | it can hold. That is why the flat regions belo |
| a temperate climate , and a rather moderate | amount of rainfall | is facing water shortages due to drought and |
| cts as a \" ; blanket\" ; to limit the | amount of radiation | the Earth loses into space. The greenhouse e |
| ing metazoan life forms began to appear. The | amount of oxygen | in the atmosphere has gone up and down dur |
| contiguous U. S. experienced nearly the same | amount of snowfall | for the month of October as the month of No |
| ed from the condition of the pipeline and the | amount of methane | flowing through. Carbon dioxide ( CO2 ) spev |

Figure 2: Sample of the results for the CQL query *amount* preceded by a noun or followed by *of* and any noun in the Meteorology subdomain.

With CQL queries, a user can also compare the frequency of different variants of multiword expressions. For example, in the term *geologic time scale, geologic* can be replaced by *geological* and *time scale* can be written as a single word. With the CQL query *[lemma="geologic.*"] ([lemma="timescale"]|([lemma="time"] [lemma="scale"]))* we can retrieve all the concordances where all the variants appear, and with the *Frequency – Node forms* feature we can see which form is more frequent (Figure 3).

| word | Frequency | Items: 9 \|\| Total frequency: 139 |
|---|---|---|
| P \| N  geologic time scale | 41 | |
| P \| N  geological time scale | 37 | |
| P \| N  geological timescale | 18 | |
| P \| N  geological timescales | 17 | |
| P \| N  geological time scales | 10 | |
| P \| N  geologic timescale | 5 | |
| P \| N  geologic time scales | 5 | |
| P \| N  geologic timescales | 3 | |
| P \| N  Geologic time scale | 3 | |

Figure 3: Frequency of variants of *geologic time scale* in the EEC.

Another feature of Sketch Engine that permits users to fully exploit the EEC is *Frequency – Text type*. With this feature, users can observe how language expression changes across different levels of expertise in the environmental domain. For instance, when searching for the verb *liquefy*, concordances can be filtered according to the user type parameter. Not surprisingly, the verb appears more often in expert-related texts than in texts addressed to the general public (Figure 4).

| User | Frequency | Rel [%] | Items: 3 \|\| Total frequency: 217 |
|---|---|---|---|
| P \| N  Expert | 153 | 125.10 | |
| P \| N  Semi-expert | 52 | 83.60 | |
| P \| N  General public | 12 | 39.10 | |

Figure 4: Frequency of *liquefy* in the EEC according to user type.

With this feature the frequency of terms in different domains can also be observed, thus verifying if a term is more specific to one domain or another. For instance, by searching the lemma *photovoltaic* and looking up its frequency according to domain, the results show that it is a term mainly linked to the domain of Renewable Energy, although it also occurs, but with much lower frequency, in Climatology and Air Quality Management (Figure 5).

| Domain | Frequency | Rel [%] | |
|---|---|---|---|
| P \| N  3.5.1 Renewable Energy | 106 | 2,683.70 | Items: 4 \|\| Total frequency: 174 |
| P \| N  2.7.2 Climatology | 34 | 104.10 | |
| P \| N  3.2.5.3 Air Quality Management | 17 | 249.40 | |
| P \| N  0 General | 17 | 85.00 | |

Figure 5: Frequency of *photovoltaic* in the EEC according to environmental subdomain.

## 3.2 Word Sketch and Sketch Diff

The EEC employs both the default sketch grammar for English underlying the word sketches in the tool in combination with the ESSG. Users can benefit from Sketch Engine's default word sketches when searching for the collocations that are used more often in specialized discourse in combination with a certain term. For instance, Figure 6 shows the modifiers of *methane*, the nouns modified by *methane* and the verbs that collocate with *methane* both as object and subject.

| modifiers of "methane" | | | nouns modified by "methane" | | | verbs with "methane" as object | | | verbs with "methane" as subject | | |
|---|---|---|---|---|---|---|---|---|---|---|---|
| | | 17.94 | | | 34.14 | | | 15.42 | | | 13.06 |
| dioxide | 35 | 9.80 | emission | 75 | 8.24 | produce | 41 | 6.58 | be | 151 | 3.45 |
| atmospheric | 27 | 6.47 | gas | 54 | 7.70 | be | 36 | 3.08 | have | 24 | 3.52 |
| coalbed | 17 | 10.49 | production | 43 | 7.40 | release | 29 | 8.35 | increase | 7 | 5.57 |
| co2 | 15 | 8.71 | hydrate | 38 | 10.60 | include | 20 | 4.85 | react | 5 | 8.23 |
| gas | 9 | 6.33 | oxide | 30 | 8.64 | emit | 14 | 7.93 | cause | 4 | 4.00 |

Figure 6: Word sketches of *methane* extracted from the EEC.

Thanks to the ESSG, users can access ready-made semantic word sketches such as those shown in Figure 7, where search terms may appear related to their hyponyms (i.e. *microorganism*), the whole they are part of (i.e. *oxygen*), their underlying causes (i.e. *tsunami*), etc.

| "microorganism" is the generic of… | | | "oxygen" is part of… | | | "tsunami" is caused by… | | |
|---|---|---|---|---|---|---|---|---|
| | | 13.42 | | | 6.36 | | | 16.05 |
| bacterium | 29 | 10.78 | atmosphere | 23 | 9.80 | earthquake | 93 | 11.41 |
| fungus | 15 | 10.49 | molecule | 21 | 10.18 | landslide | 43 | 10.71 |
| pathogen | 5 | 9.35 | compound | 18 | 9.51 | eruption | 26 | 9.79 |
| alga | 5 | 8.42 | water | 16 | 8.44 | water | 22 | 8.14 |
| virus | 4 | 8.76 | earth | 15 | 9.26 | slide | 13 | 9.14 |

Figure 7: Semantic word sketches of *methane* extracted from the EEC.

The word sketch queries can be complemented with the text type filters provided by the tags of the EEC (or subcorpora based on them). In this sense, users can also observe how concepts can change their relational behavior across different environmental subdomains. For example, Figure 8 shows how *nitrogen* is mainly categorized as a type of *pollutant* in the domain of Air Quality Management and as a type of *nutrient* in that of Biology.

| "nitrogen" is a type of… | | |
|---|---|---|
| | | 8.06 |
| pollutant | 9 | 8.61 |
| gas | 5 | 6.27 |

| "nitrogen" is a type of… | | |
|---|---|---|
| | | 3.27 |
| nutrient | 4 | 8.55 |
| gas | 3 | 5.56 |

Figure 8: *Nitrogen* generic-specific semantic word sketches in Air Quality Management (left) and Biology (right) subcorpora.

Additionally, if users access the concordances extracted with the ESSG, they can extract knowledge-rich contexts (i.e. contexts containing domain knowledge potentially useful for conceptual analysis (Meyer 2001)) like the ones in Table 1.

Table 1: Sample of knowledge-rich contexts extracted from the EEC with the aid of the ESSG.

| | |
|---|---|
| generic-specific | A <u>hydrograph</u> is a <u>graph</u> that reflects the discharge of a river over a period of time.<br>The <u>astronomical tide</u> refers to the <u>regular oscillations of the sea or ocean surface</u>[…]. |
| part-whole | <u>Sand grains</u> usually consist of <u>quartz</u> but may also be fragments of feldspar, mica, and, […].<br><u>Seawater</u> contains <u>sodium chloride</u> and other salts in concentrations three times greater […]. |
| location | <u>Lagoons</u> commonly form on <u>coastlines</u> that are subsiding, or where sea level is rising.<br>Most <u>ozone</u> is found in the <u>stratosphere</u> at elevations between 10 and 50 kilometers […]. |
| cause | […] the human costs of malaria outweigh the <u>environmental damage</u> caused by the <u>use of DDT</u>.<br><u>Logging</u> may also contribute to <u>deforestation</u> by making it easier for agriculture to […]. |
| function | <u>Membrane-assisted BAC</u> is used for the <u>removal of priority pollutants</u> from secondary […].<br><u>Liquid-in-glass thermometers</u> are often used for <u>measuring surface air temperature</u> because […]. |

Another word-sketch based feature that can be especially exploited with the EEC is *Sketch diff*. It allows the user to compare either the word sketches of two lemmas, the word sketches on the same lemma in two subcorpora, or two different word forms of the same lemma. Figure 9 shows an example of each type. At the left, the modifiers of *risk* (in green) and *hazard* (in red) in the whole EEC are contrasted. As it can be observed, these two semantically related terms tend to co-occur with different modifiers, although they also share some of them (in white). At the center, there is a sketch diff that shows how *water* takes different verbs as an object in Hydrology (in green) and Water Treatment and

| modifiers of "risk/hazard" | 2,393 | 2,099 | 0.36 | 0.70 |
|---|---|---|---|---|
| extinction | 30 | 0 | 8.4 | -- |
| cancer | 27 | 0 | 8.4 | -- |
| disaster | 30 | 0 | 8.3 | -- |
| disease | 21 | 0 | 7.7 | -- |
| great | 87 | 15 | 8.1 | 5.6 |
| flood | 180 | 39 | 9.8 | 7.7 |
| erosion | 106 | 24 | 9.0 | 6.9 |
| flooding | 23 | 5 | 8.1 | 6.0 |
| drought | 25 | 7 | 7.9 | 6.2 |
| potential | 99 | 49 | 8.6 | 7.6 |
| health | 142 | 81 | 10.1 | 9.4 |
| serious | 20 | 35 | 7.5 | 8.4 |
| tsunami | 18 | 53 | 7.1 | 8.8 |
| safety | 6 | 31 | 5.8 | 8.3 |
| climate-related | 4 | 37 | 5.7 | 9.1 |
| geologic | 4 | 42 | 5.1 | 8.6 |
| earthquake | 1 | 16 | 3.5 | 7.6 |
| natural | 7 | 290 | 4.1 | 9.5 |
| aviation | 0 | 13 | -- | 7.5 |
| geological | 0 | 37 | -- | 8.3 |

| verbs with "water" as object | 1,003 | 1,304 | 0.15 | 0.14 |
|---|---|---|---|---|
| infiltrate | 8 | 0 | 7.9 | -- |
| entrain | 9 | 0 | 7.9 | -- |
| ground | 8 | 0 | 7.8 | -- |
| flow | 15 | 1 | 8.4 | 4.2 |
| pour | 7 | 2 | 7.7 | 5.5 |
| evaporate | 13 | 8 | 8.4 | 7.4 |
| divert | 11 | 10 | 8.2 | 7.7 |
| contaminate | 8 | 8 | 7.6 | 7.3 |
| withdraw | 7 | 8 | 7.6 | 7.5 |
| pump | 15 | 17 | 8.1 | 8.1 |
| disinfect | 6 | 9 | 7.5 | 7.8 |
| treat | 25 | 55 | 8.4 | 9.4 |
| supply | 9 | 24 | 7.1 | 8.3 |
| drink | 35 | 134 | 9.5 | 11.1 |
| receive | 10 | 37 | 6.5 | 8.3 |
| purify | 1 | 8 | 4.9 | 7.5 |
| produce | 13 | 117 | 4.9 | 8.0 |
| conserve | 1 | 14 | 4.3 | 7.9 |
| intend | 1 | 44 | 4.4 | 9.6 |
| regulate | 0 | 18 | -- | 8.1 |

| verbs with "gas" as subject | 983 | 877 | 0.10 | 0.20 |
|---|---|---|---|---|
| clean | 16 | 0 | 8.9 | -- |
| stage | 8 | 0 | 8.0 | -- |
| fire | 7 | 0 | 7.8 | -- |
| exit | 6 | 0 | 7.5 | -- |
| enter | 11 | 0 | 7.3 | -- |
| reheat | 4 | 0 | 7.1 | -- |
| flare | 4 | 0 | 7.0 | -- |
| mix | 15 | 1 | 7.8 | 4.0 |
| play | 14 | 2 | 7.5 | 4.8 |
| exert | 12 | 4 | 8.1 | 6.6 |
| leave | 4 | 6 | 6.1 | 6.8 |
| build | 3 | 5 | 5.9 | 6.7 |
| pass | 4 | 7 | 5.8 | 6.7 |
| contribute | 7 | 13 | 6.4 | 7.4 |
| escape | 2 | 5 | 5.9 | 7.3 |
| absorb | 11 | 30 | 7.6 | 9.1 |
| expand | 2 | 8 | 5.4 | 7.5 |
| radiate | 0 | 4 | -- | 6.9 |
| diffuse | 0 | 5 | -- | 7.4 |
| trap | 0 | 10 | -- | 8.2 |

Figure 9: Sample of sketch diffs extracted from EEC.

Supply (in red), as well as a considerable number of shared results. Finally, the sketch diff at the right outlines the verbs that tend to have *gas* as subject in singular (in green) and in plural (in red) in the whole EEC.

### 3.3   Word List

The *Word list* feature can be used to extract frequency lists with many different settings including n-gram extraction, filtering based on regular expressions or keyword extraction with the aid of a user-chosen reference corpus. This feature can be used in combination with an EEC subcorpus, which allows the user to generate very specific frequency lists. Some examples of frequency lists that could be useful to generate from the EEC are: nouns specific to Energy Engineering academic texts using the British National Corpus as a reference; most common 4-grams in Zoology texts; adjectives containing *-friendly* in the whole EEC; or the most common verbs in Geology texts (Figure 10).

| lemma (lowercase) | Frequency |
|---|---|
| form | 631 |
| see | 251 |
| cause | 231 |
| occur | 230 |
| move | 216 |
| become | 208 |
| determine | 188 |
| represent | 175 |
| produce | 167 |
| rise | 166 |
| make | 154 |
| provide | 142 |
| include | 124 |
| take | 108 |
| show | 108 |
| change | 108 |
| grow | 105 |
| develop | 105 |
| increase | 104 |
| reach | 98 |
| flow | 98 |

Items: 1,195 || Total frequency: 14,845

Figure 10: Frequency list of verbs in Geology texts.

## 4   Conclusion

In this paper, we have shown how the EEC was built and compiled and how it can be queried and exploited in Sketch Engine. The EEC's metadata, the default sketch grammar and the ESSG make the EEC a useful resource for any user interested in environmental science. As future work, we will refine, improve and update the ESSG and develop new rules for Spanish. Furthermore, in the short term, we plan to upload an improved version of the EEC (with more words and some minor codification issues solved) and a first version of the Spanish counterpart. In the long run, we will enhance the EEC with a new annotated version, where different semantic tags will be added to improve its querying potential. These semantic tags will include semantic categories and argument structure.

Sketch Engine's API also allows for the exploitation of the EEC from external applications. An example of this is EcoLexiCAT, a terminology-enhanced computer assisted translation (CAT) tool that provides easy access to domain-specific terminological knowledge in context (León-Araúz & Reimerink, 2018; León-Araúz, Reimerink & Faber, 2017). EcoLexiCAT integrates different features of the professional translation workflow in a stand-alone interface where a source text is interactively enriched with terminological information (i.e., definitions, translations, images, compound terms, corpus access, etc.) from EcoLexicon, BabelNet, IATE, and Sketch Engine. In the Sketch Engine module of EcoLexiCAT's interface, terms from both the source and target segments can be selected and direct access is given to concordances, CQL queries and word sketches of the selected terms. For a more detailed analysis, the output of the queries can be opened in a new tab that sends users to the website of the Sketch Engine Open Corpora.

**Acknowledgements**

This research was carried out as part of projects FF2014-52740-P, Cognitive and Neurological Bases for Terminology-enhanced Translation (CONTENT), and FFI2017-89127-P, Translation-oriented Terminology Tools for Environmental Texts (TOTEM), funded by the Spanish Ministry of Economy and Competitiveness.